\documentclass{article}

\usepackage{arxiv}

\usepackage[utf8]{inputenc} 
\usepackage[T1]{fontenc}    
\usepackage{hyperref}       
\usepackage{url}            
\usepackage{booktabs}       
\usepackage{amsfonts}       
\usepackage[numbers]{natbib} 
\usepackage{nicefrac}       
\usepackage{microtype}      
\usepackage{lipsum}		
\usepackage{graphicx}
\usepackage{natbib}
\usepackage{doi}

\title{Comparative Analysis of Contextual Relation
Extraction based on Deep Learning Models}

\author{ { R. Priyadharshini*} \\
	Department of Computer Science\\
	  Pondicherry University\\
	Puducherry, India \\
	\texttt{priyadharsini.r02@gmail.com} \\
	\And
	{ G. Jeyakodi} \\
	Department of Computer Science\\
	  Pondicherry University\\
	Puducherry, India \\
	\texttt{rjeyakodi02@gmail.com} \\
 \And
	{ P. Shanthi Bala} \\
	Department of Computer Science\\
	  Pondicherry University\\
	Puducherry, India \\
	\texttt{shanthibala.cs@gmail.com} \\
}

\renewcommand{\shorttitle}{\textit Comparative Analysis of Contextual Relation
Extraction based on Deep Learning Models}

\hypersetup{
pdftitle={A template for the arxiv style},
pdfsubject={q-bio.NC, q-bio.QM},
pdfauthor={David S.~Hippocampus, Elias D.~Striatum},
pdfkeywords={First keyword, Second keyword, More},
}

\begin{document}
\maketitle

\begin{abstract}

Contextual Relation Extraction (CRE) is mainly used for constructing a knowledge graph with a help of ontology. It performs various tasks such as semantic search, query answering, and textual entailment. Relation extraction identifies the entities from raw texts and the relations among them. An efficient and accurate CRE system is essential for creating domain knowledge in the biomedical industry. Existing Machine Learning and Natural Language Processing (NLP) techniques are not suitable to predict complex relations from sentences that consist of more than two relations and unspecified entities efficiently. In this work, deep learning techniques have been used to identify the appropriate semantic relation based on the context from multiple sentences. Even though various machine learning models have been used for relation extraction, they provide better results only for binary relations, i.e., relations occurred exactly between the two entities in a sentence. Machine learning models are not suited for complex sentences that consist of the words that have various meanings. To address these issues, hybrid deep learning models have been used to extract the relations from complex sentence effectively. This paper explores the analysis of various deep learning models that are used for relation extraction.
\end{abstract} 

\section{Introduction}
Contextual Relation Extraction (CRE) helps to understand the meaning of the entities and their relationship in a sentence. It can improve the performance of Natural Language Processing tasks such as information retrieval, question answering, and semantic search [1]. Named Entity Recognition aims to automatically identify and classify objects like people, products, organizations, locations, etc. The process of identifying the terms in a text and arranging in an appropriate group is a source for named entity recognition and a key component for text analysis. The analysis of common syntactic patterns is an important factor of NER. Many deep learning models solve entity recognition applications such as indexing documents, finding relationship among entities, and building an ontology [2-4]. The combination of NER and CRE can provide a rich understanding of the text by identifying both the entities and their relationships based on the context. The joint modeling of entity recognition and relation classification attained more focus recently [5]. Additionally, these end-to-end models have generated massively to improve the results. Information Extraction (IE) begins with the creation of knowledge graphs that transforms unformatted text into formatted data. Entity extraction and Relation extraction are the two subtasks of IE. \\
Relation extraction is ongoing research for the recent years. Neural networks enabled technology is used to efficiently classify entities and relation. Natural Language Understanding (NLU) represents the associated relationship among the existing objects and a distinct relationship between two or more entities. Entity relationship is the basis for automatically creating a knowledge graph. Relation extraction instantly detect and categorizes the entities from the text during semantic relationship extraction. Example of binary and n-ary relation are shown in Figure.1.

\begin{figure}
	\centering
		\includegraphics[width=0.45\textwidth]{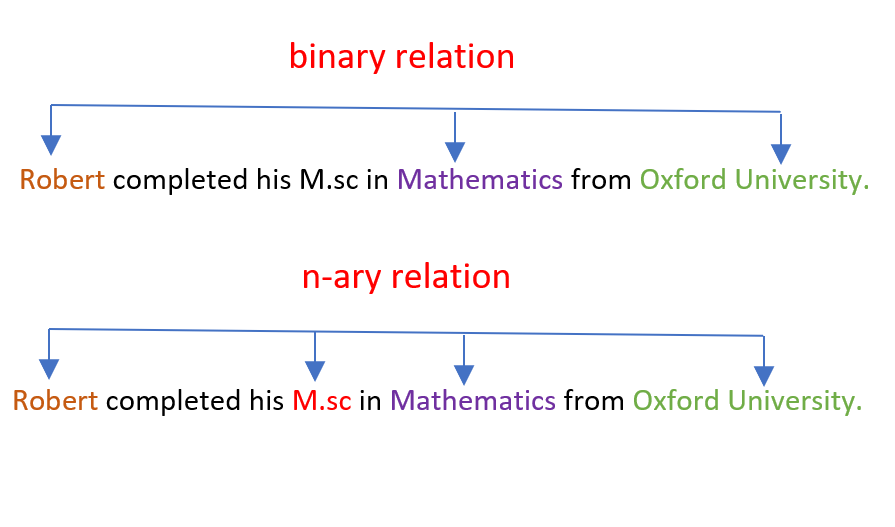}
  \caption{ Example of binary and n-ary relation.}
\end{figure}
Binary relation consists of two entities and one relation and n-ary relation consists of more that two entities and many relations. Binary relation extraction models may have trouble in handling larger sentence and take lot of time for processing. Some of the common issues in binary relation extraction are ambiguity, incomplete data and noise in the text. To find and understand the connections between different established categories, RE makes use of a range of technologies. Recent joint extraction models work on fixed word vector format for word embedding that are unsuitable for a word that has multiple semantic meanings. To address this problem, Bo Qiao et al. developed a dynamic fine-tuning method to overcome the issues in static word embedding using the LSTM-LSTM-Bias method proposed by Zheng et al [6]. \\
Bidirectional Encoder Representations from Transformers (BERT), is a machine language pre- training model to represent language. BERT uses joint conditions to compare each word context in forward and backward directions. The BERT model can be improved by adding a single additional output layer for tasks such as question answering and language inference. It does not require major changes in the architecture. Devlin et al. proposed the significance of bidirectional pre-training for language representations to eliminate the requirement of multiple task-specific architectures. The BERT model is based on the fine tune representation that outperforms multiple task-specific architectures and reaches cutting-edge performance on a variety of task levels, including token and sentence levels. The pre-training and fine-tuning steps in BERT architecture help to understand the semantic meaning of the words effectively. Before solving the joint 
extraction task, it pre-trains the BERT model using another corpus. BERT can be used for a wide range of linguistic activities and primarily adds a thin layer to the basic model [7]. Figure 2 shows the categorization of various BERT (Fine Tuning) based applications. \\
\begin{figure}
	\centering
		\includegraphics[width=0.45\textwidth]{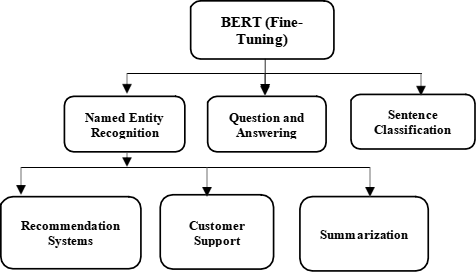}
 \caption{ BERT (Fine Tuning) based applications.}
\end{figure}
\section{RELATED WORK}
In this section, various models for Relation Extraction (RE) are explored. Relation extraction is used to understand the relationships among the various entities in an unlabeled text. There are various methods to perform relation extraction, from a simple string extraction to automated models.
\subsection{Models for Relation Extraction}
Recently, many works such as document-level, pipelined and joint model is proposed to solve the Relation Extraction tasks.
\begin{itemize}
\item  Pipelined Method: The pipeline method treats NER and Relation Categorization as a distinct operation. Zexuan et al. suggested the new state-of-the-art for entity and RE using a straightforward pipelined strategy, and they obtain a relative improvement over the earlier joint models using a similar pre-trained encoder [8].
\item Joint model: Joint extraction model recognizes entities and relations simultaneously and these models extract entities and relations using a single task. Feature-based structured systems compose the majority of joint techniques. Zheng et al suggested a tagging scheme to convert joint extraction of entities and relations [9].
\item Document-level Relation Extraction Models: When compared to sentence-level Relation Extraction, document level Relation Extraction is a complex process. Because document may contain entity pairs with multiple relationships. The Sentence Importance Estimation and Focusing (SIEF) framework was presented for document-level. In various disciplines, the SIEF framework enhances the performance of basic models [10]. Zeng et al. proposed an architecture to distinguish the document level based on the intra and inter sequential reasoning techniques [11].
\end{itemize}
\subsection {Contextual Word Embeddings for Relation Extraction}
Word embeddings are a method for finding similarities between words in a corpus by predicting the co-occurrence of words in a text using some sort of model. When it was proven that word embeddings could be used to find analogies, they became well-known in the field of automated text analysis. Table 1 illustrates various word embedding techniques.
\begin{table}
\caption{Word Embedding Techniques}
\centering
\begin{tabular}{|p{1cm}|p{3cm}|p{4cm}|p{4cm}| }
\toprule          
\cmidrule(r){1-2}
S. No & Word Embeddings & Explanation & Feature \\
\midrule
1 & TF-IDF  & A statistical technique for determining a word's relevance to the corpus of text. It doesn't record word associations with semantic meaning. & Perform well on retrieving information and extracting keywords from documents.   \\
2 & Word2Vec & CBOW and Skip-gram architectures based on neural networks are superior at
capturing	semantic information. & Suitable for smaller and larger datasets. \\
3 & GLoVe & Global word-word co- occurrence-based matrix factorization. It resolves Word2Vec's local context issues. & Better in tasks that involve word analogies and named-entity recognition. Word2Vec is commonly used in semantic analysis tasks. \\
4 & BERT & High-quality contextual information can be captured via a transformer-based attention method. & Translation services and a question-and-answer platform are used in the Google Search engine to interpret search keywords. \\
\bottomrule
\end{tabular}
\label{tab:table1}
\end{table}
Contextual embeddings represent each word based on its context, capture the word usage across a range of situations and encode cross-linguistic knowledge. Contextual embeddings, such as ELMo and BERT, perform significantly better than generic word representations. The ELMo of the bidirectional Language Model combines the representations from its intermediary layer according to the task at hand. When ELMo combines both the representations of forward and backward LSTMs, the interactions between the left and right contexts are not taken into consideration [12]. BERT offers Masked Language Modeling (MLM) that involves randomly masking some of the tokens in input sequence. It employs a Transformer encoder during pre- training to focus on instances involving bi-directional communication and the other one is Next Sentence Prediction (NSP). RE with Distant Supervision and Transformers suggested by Despina et al. predicts better embeddings using fine-tuning BERT [13]. ELMo and BERT perform better than Word2Vec, and offer ground-breaking performance in a range of NLP applications. Using two input sentences, natural language processing (NLP) determines whether the preceding sentence follows the first one. NLP helps to facilitate the tasks which needs sentence pairs analysis.

\subsection{Datasets for Relation Extraction}
Several datasets for relation extraction have been developed recently to enhance the relation extraction systems. Two examples of RE datasets created through human annotations with relation types are SemEval-2010 Task 8 and ACE05. The crowdsourcing method is used to build TACRED dataset to meet the demands of the large-scale dataset. To enhance document-level RE research, DocRED was developed. Ten thousand annotated examples and more than one hundred relations are included in FewRel. The issues with few-short relation extraction have been addressed with the development of FewRel and FewRel 2.0. HacRED consists of 65,225 relational facts that has been identified from 9,231 documents [14-18].

\section{ANALYSIS OF DEEP LEARNING MODELS}
Deep learning uses artificial neural networks using representation learning. It can be supervised, semi-supervised, or unsupervised. The rapid growth and use of Artificial Intelligence based systems have elevated concerns regarding understandability [19]. Rahman et al. constructed artificial neural networks model for effectively forecast solar radiation [20]. Representation learning helps to reduce the data dimension to simplify in identifying patterns and anomalies. A neural network instructs computers to scrutinize data like human brain. The hidden layers are referred as the term "deep." Deep neural networks consist of 150 hidden layers, compared to the two or three layers that traditional neural networks normally have. The structure of deep neural network is depicted in Figure 3.
\begin{figure}
	\centering
		\includegraphics[width=0.48\textwidth]{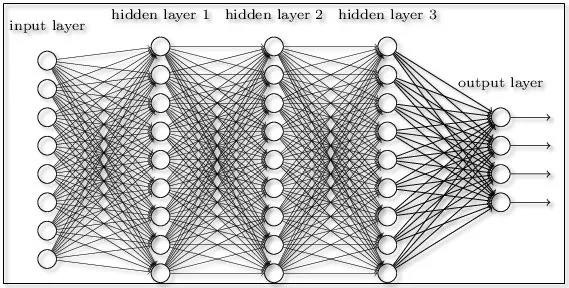}
 \caption{ Structure of Deep Neural Network.}
\end{figure}
Deep learning model helps to learn categorization that take input from the various sources such as images, text, and sounds. It can also attain high accuracy, occasionally even superior human performance. Large labeled data and multi-layered architectures are used to train the models to learn data characteristics automatically. Deep learning has the ability to achieve high levels of accuracy when trained on huge amounts of data. There are many complex problems to solve in natural language. In some specific natural language problems, deep learning achieves the best results. Table. 2 illustrates some of the deep learning techniques that are widely used in the task of RE. Survey on existing RE model on Deep learning Techniques using various dataset are mentioned in Table. 3.
\begin{table}
	\caption {Deep Learning Techniques}
	\centering
	\begin{tabular} {|p{1cm}|p{3cm}|p{3cm}|p{3cm}|p{3cm}|p{3cm}}
		\toprule          
		\cmidrule(r){1-2}
S. No     & Models    & Working Principle	 & Benifits & Issues \\
		\midrule
1 &	CNN [21-23]	& Convolutional Neural Network has Multiple layers to process and extract features.	& human supervision is not required	for features recognizing. &	Overfitting, exploding gradient, and class imbalance. \\
2 &	Bi-GRU [24]	& Model that combines the Gated Recurrent Unit (GRU) and the bidirectional	Recurrent	Neural Network. &	Simple	than LSTM	& Only	input and	forget gate.\\
3 &	LSTM [25]&	Long Short-Term Memory picks up and remembers enduring addictions and long-term retention of past knowledge.	&Offers parameters like learning rates, input, and output
biases.	&Overfitting \\
4	&  CRF [26] &	It's a discriminative model to predict contextual information.	& Perform well on NLP tasks such as part of speech tagging, NER. & More accurate but difficult	to train. \\
5 & BiLSTM [27]	& It is a combination of two separate RNNs.The networks access both forward and backward information. &	Better predictions compared to Auto Regressive Integrated Moving Average
(ARIMA). &	Slower	and requires more time. \\
6 &	RNN [28]	&RNN has connections that form directed cycles that allows the current phase to accept the LSTM outputs as inputs. 	&Remembers every piece	of information through time.& Exploding gradient problem, and long-term dependency of words. \\
7 & MLPs [29]	& Made up of many layers of perceptron with activation capabilities. Layers of input and output are interconnected and have equal layers for the input and output. &	Used to solve complex nonlinear problems. &	Feature scaling, and Computational complexity. \\
8 &	DBNs [30] &	Made up of a lot of latent and random layers. Latent variables, often called hidden units that are characterized by binary values. Boltzmann machines has connections between its layers.	& Powerful and learn complex patterns. Process large amounts of data very quickly.& Hardware requirements, expensive to train. \\
9 &	RBM [31]	& Consists of both visible and hidden components. All hidden units are linked to all visible units.	& Computationally efficient and faster than	a	typical Boltzmann Machine.	& Hard	to evaluate	or simulate. \\
	\bottomrule
	\end{tabular}
	\label{tab:table2}
\end{table}
\begin{table}
	\caption {Comparison of Deep Learning Relation Extraction Models}
	\centering
	\begin{tabular}{|p{1cm}|p{2.5cm}|p{3cm}|p{3cm}|p{3cm}|p{2.5cm}|}
		\toprule          
		\cmidrule(r){1-2}
S. No &	Authors and Year &	Objective &	Techniques &	Issues &Dataset \\
		\midrule
1 & Chen Gao et al 2022 [32] & It extracts the semantic mutuality between entity and relation extraction. & HBT, WDec and CasREL	& Overlapping entities in the sentence cannot be resolved by this technique. & New	York Times (NYT), WebNLG \\
2 & O.A Tarasova et al 2022 [33] & Method to extract Clinical Named Entities from texts which combines the naive	Bayes classifier with specially	built filters.&	Naïve Bayes classifier & The result of CNER	using naive-Bayes method is slightly worse. & CHEMDNE R \\
3& T.Bai et al 2022 [34]& Segment attention method based on CNN to extract local semantic
properties through word embedding.	&	SVM, KNN, CNN,	and SEGATT- CNN	&	This	model applies only to supervised methods	&	Herbal- Disease and Herbal Chemistry, HD-HC \\
4& Qingbang W et al 2022 [35]&	This model efficiently predicts the information and
semantic context of the current text. &BERT- BiLSTM, BiLSTM- ATT &The BERT- BLSTM
network does not function well	when dealing with the issue of partial entity overlap.	&	Food public opinion field data \\
5& Hailin Wang et al 2022 [36]	&	Supervised and distant supervision methods	for Relation Extraction.& DNN, RNN and PCNN	&	Error propagation	in supervised methods.	&	SemEval 2010-task8, ACE	series and NYT+Freebase \\
6	&	Yang Yang et al 2022 [37]	&	Basics of IE and DL, mainly concentrating on DL technologies in the field of IE.& RNN, CNN and BiLSTM&		DNN	models cannot	handle all		the
knowledge	in huge database.&		COVID-19 news \\
7 & Zhiyun Z et al 2022 [38]&	Distant Supervised Relation Extraction (DSRE) model using residual network.&	CNN-ATT, PCNN-ATT,	and DSRE	&Noise	label reducing.	&Freebase	+ NYT \\
8	&Guangyao Wang et al 2022 [39]	&Weighted		graph convolutional network	(WGCN) model to extract the nontaxonomic relationships.&	LSTM, CNN	and BiLSTM	&When	the feature graph is used as an input to the GCN, the directed graph's effect is not better.&	Human- annotated RE data,	NYT data \\
9	&Chantrapor nchai et al 2021 [40]&	BERT and spacy model to extract specific information from entire texts based on machine learning.	&BERT, Spacy	&The Performance of SpaCy is poor. &Tourism Data \\
10	&W. Zhou et al 2021 [41]&	The multi-label and multi-entity problems are solved using	adaptive thresholding	and localized	context pooling.&	BERT- ATLOP, BERT-E	&Adaptive thresholding only	works when the model is optimized.	&DocRED, CDR \\
\bottomrule
	\end{tabular}
	\label{tab:table3}
\end{table}

\begin{table}
		\centering
	\begin{tabular}{|p{1cm}|p{2.5cm}|p{3cm}|p{3cm}|p{3cm}|p{2.5cm}|}
		\toprule          
		\cmidrule(r){1-2}

11	&Prashant S et al 2021 [42]	&Attention Retrieval Model to
improve	the applicability of attention-based model for RE.&	LSTM, RNN,	and GRU	&The ARM technique must test the model rather than categorize the text.&Atlas	of Inflammation Resolution (AIR), BioGRID,
and ChemProt. \\
12	&Liu Kang et al2020 [43]&	Neural relation extraction with a specific focus to train neural relation extraction model.&	BERT, LSTM, and BiLSTM	&Unable to meet demand	in practical applications.&	ACE, SemEval 2010, TACRE.	\\
13	&Boran Hao et al 2020 [44]&	Novel joint training technique is used to develop language model pre-training for clinical corpora.	&Clinical BERT	+ BiLSTM, Clinical KB- ALBERT&	The improvement for ALBERT is less significant.	&MIMIC-III and	UMLS Knowledge Base \\
14&	Diana Sousa et al 2020 [45]	&BiOnt	uses	four different kinds of biomedical ontologies	to perform	relation extraction. & BO-LSTM, BioBERT &	This approach does not allow for th integration of ontological knowledge.	& DDI corpus, PGR corpus, BC5CDR corpus \\
15 & Rakesh Patra et al 2019 [46] & A Model for automatic generation of named entity distractor.	A combination of statistical and semantic similarity is used. & F requency based,
Co- occurrence based	& Existing techniques focus on language learning and vocabulary testing,	these metrics are not applicable for evaluating	the named entity distractors.	& 200 cricket -related MCQ-key pairs \\
16	& Veera Ragavendra et al 2018 [47]	& The rule-based method	for relationship classes. &SVM, BiLSTM&	Suitable only for smaller number of samples. &I2b2 2010 \\
17&	S. Zeng et al 2018 [48]	& Separate Intra- and Inter-sentential Reasoning		for Document-level Relation Extraction (SIRE) architecture.	&BiLSTM,BERT, and SIRE- BERT&	This model primarily enhances intra- sentential relations' performance.	&DocRED, CDR, and GDA \\
18 &	Jing Qiu et al 2018 [49]	&SGNRI model to extract	non- taxonomic associations using a multi-phase correlation	search automated system.	&SGNRI(Word2Vec) SGNRI(LD A)	&The performance of the	word2Vec- based model is poor.	&Concept Pairs \\
19&	Henghui et al 2018 [50]&	Developed a model for the purpose of clinical feature extraction using a contextual word embedding approach.	&BiLSTM- CRF, ELMo	&Difficulties    in
creating	a language model on a big corpus of domain- specific data.	&I2B2 2010 \\
20&	Linfeng Song et al 2018 [51]	&Graph-state LSTM model	for displaying discourse	and relationship structures.&	Bidirection al	DAG LSTM, GLSTM	&Word	sense confusion&	Biomedical domain \\
\bottomrule
	\end{tabular}
	\label{tab:table4}
\end{table}

\section{DISCUSSION}
The comparison of existing relation models with various techniques shows that BERT based relation extraction model provides significantly improved performance than other models such as CNN, RNN, KNN, etc. BERT reads text input in both left-to-right and right-to-left directions at once. Using this bidirectional capability, BERT is pretrained on two different NLP tasks such as Masked Language Modeling and Next Sentence Prediction. It is observed that the model can be used for various domains such as clinical, tourism, agriculture, and so on. Table 4 shows the performance evaluation of existing relation extraction models based on deep learning techniques. Table 5 lists the performance accuracy (F1 score) of the BERT, CNN, and RNN based models for the SemEval 2010 dataset. 
\begin{table}
\caption { Performance evaluation of existing relation extraction models}
		\centering
	\begin{tabular}{llll}
		\toprule          
		\cmidrule(r){1-2}
Model &	Dataset & F1 score & Reference \\
SIRE-BERT &	DocRED & 62.05 & [9] \\
RoBERT-ATLOP & DocRED &	63.40&	[41] \\
MDL-J3E &	COVID-19 News&	70.96&	[28] \\
KNN&	CDR	&71.49&	[34] \\
BiDAG LSTM &Biomedical&	75.6	&[51] \\
BERT	&Tourism data&	77.96&	[40] \\
SGNRI& Concept pairs&	81.4	&[49] \\
KB-BERT&	I2b2 2010	&84.4	&[44] \\
BiLSTM &	SemEval 2010	&84.7&	[6] \\
WCGN	&NYT Data	&84.47&	[27] \\
BERT-BiLSTM	&Food public opinion field data&	87.44	&[35] \\
ELMo+BiLSTM-CRF&	I2b2/VA 2010	&88.60&	[50] \\
BERT-BiLSTM-CRF	&Clinical data	&96.73	&[52] \\
\bottomrule
	\end{tabular}
	\label{tab:table5}
\end{table}

\begin{table}
\caption {Comparison of CNN, RNN, and BERT based RE models (SemEval 2010 dataset)}
		\centering
	\begin{tabular}{lll}
		\toprule          
		\cmidrule(r){1-2}
Model&	F1 Score	&Reference \\
Att+CNN	&88.0	&[53] \\
CR-CNN	&84.1	&[54] \\
MVRNN	&82.4	&[55] \\
BRCNN	&85.4	&[56] \\
Att+BiLSTM&	84.0  &	[57] \\
R-BERT&	89.2	&[58] \\ 
BERT-GCN&	90.2&	[59] \\
\bottomrule
	\end{tabular}
	\label{tab:table6}
\end{table}
The F1 statistical metric is employed to calculate an estimation of the deep learning model’s accuracy. From the literature survey, it has been identified that the BERT-BiLSTM-CRF model achieves better results for breast cancer concepts and their attributes extraction. Even though several BERT based relation extraction for different fields are developed, the overlapping of relation and partial entity overlapping are still in a development state. \\
Relation extraction offers a wide range of applications including information retrieval, question answering, and knowledge base construction, etc. Creating models that can extract relationships in a multilingual and cross-lingual situation is another important area of focus. Additionally, the combination of relation extraction with other NLP tasks such as named entity recognition and event extraction is expected to lead to more wide-ranging and sophisticated NLP systems. The contexts such as syntax, pragmatics can be considered for improving the relation prediction accuracy. BERT variants such as RoBERTa, DistilBERT, and XLNet can be incorporated to enhance the contextual relation prediction.
\section{CONCLUSION}
This paper provides information on conceptual relation extraction and the various techniques used. It affords information on various deep learning models which are used in different tasks such as building classification models, developing recommendation systems, learning behavior predictions, and so on. It has been identified that BERT- based models can provide better accuracy to identify relations based on their context from multiple sentences. While comparing to other models BERT- BiLSTM-CRF achieved 97\% of accuracy with limited information. In future, the overlapping relations problems can be focused to improve prediction accuracy.

\end{document}